\documentclass[letterpaper, 10 pt, conference]{ieeeconf}  

\IEEEoverridecommandlockouts                              

\usepackage{amsmath} 
\usepackage{amssymb}  
\usepackage{booktabs}
\usepackage{multirow}
\usepackage{graphicx}
\usepackage{subcaption}
\usepackage{xcolor}
\usepackage{makecell}

\title{\LARGE \bf
PASTA: Vision Transformer Patch Aggregation for Weakly Supervised Target and Anomaly Segmentation
}

\author{Melanie Neubauer$^{1}$, Elmar Rueckert$^{1}$  and Christian Rauch$^{1}$  
\thanks{$^{1}$ Chair of Cyber Physical Sytems,        Technical University of Leoben, 8700 Leoben, Austria        {\tt\small melanie.neubauer@unileoben.ac.at}}%
        }

\begin{document}

\makeatletter
\long\def\@maketitle{%
  \newpage
  \begin{center}
    {\LARGE \@title \par}
    \vskip 1.5em
    {\large \lineskip .5em \begin{tabular}[t]{c}\@author \end{tabular}\par}
    \vskip 1em
    \includegraphics[width=\textwidth]{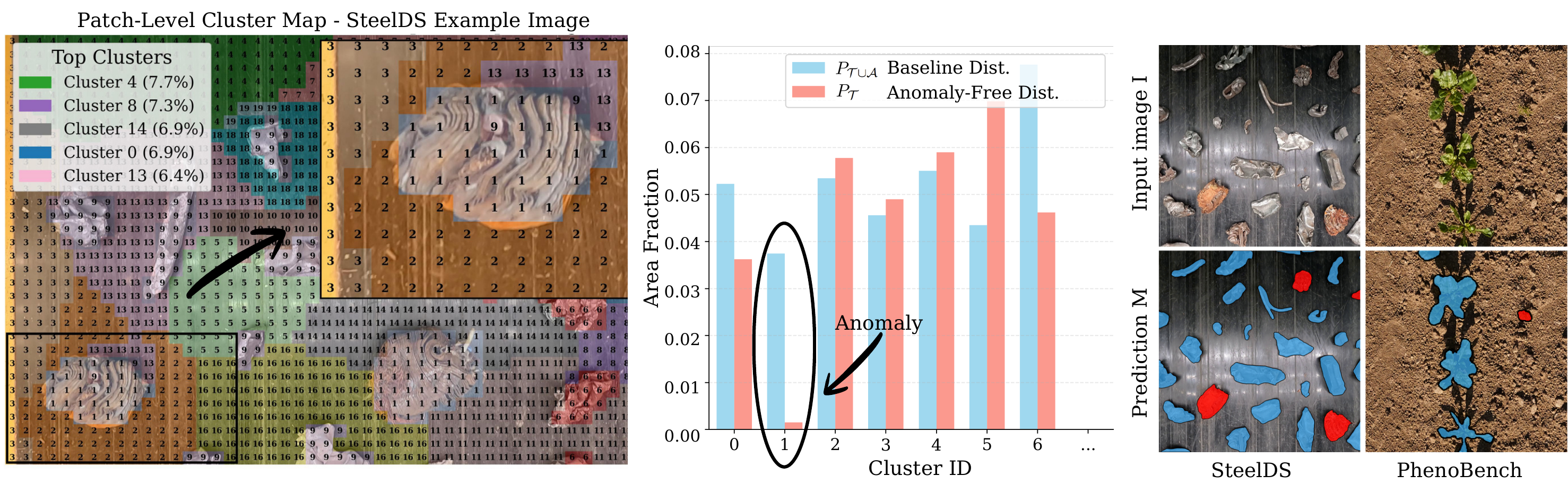}
    \def\@captype{figure}
    \captionof{figure}{
    Anomaly detection via distribution analysis. 
    Left: Patch-Level Cluster Map on a SteelDS example, highlighting the 'Top Clusters' with numbers corresponding to the defined Cluster ID. 
    Middle: Comparison of the Area Fraction between the Baseline Distribution ($P_{\mathcal{T}\cup\mathcal{A}}$, blue bars) and the Anomaly-Free Distribution ($P_{\mathcal{T}}$, orange bars), where an Anomaly is identified by significant distribution contrasts (e.g., peak at Cluster ID 1).  
    Right: Original input images alongside the final segmentation masks (blue: Targets, red: Anomalies) for both the SteelDS and PhenoBench datasets.}

    \label{fig:header}
    \vspace{-0.5cm}
  \end{center}
}
\makeatother
\maketitle


\begin{abstract}

Detecting unseen anomalies in unstructured environments presents a critical challenge for industrial and agricultural applications such as material recycling and weeding.
Existing perception systems frequently fail to satisfy the strict operational requirements of these domains, specifically real-time processing, pixel-level segmentation precision, and robust accuracy, due to their reliance on exhaustively annotated datasets.
To address these limitations, we propose a weakly supervised pipeline for object segmentation and classification using weak image-level supervision called \textit{Patch Aggregation for Segmentation of Targets and Anomalies} (PASTA).
By comparing an observed scene with a nominal reference, PASTA identifies Target and Anomaly objects through distribution analysis in self-supervised Vision Transformer (ViT) feature spaces.
Our pipeline utilizes semantic text-prompts via the Segment Anything Model 3 to guide zero-shot object segmentation.

Evaluations on a custom steel scrap recycling dataset and a plant dataset demonstrate a 75.8\% training time reduction of our approach to domain-specific baselines.
While being domain-agnostic, our method achieves superior Target (up to 88.3\% IoU) and Anomaly (up to 63.5\% IoU) segmentation performance in the industrial and agricultural domain.

\end{abstract}

\section{Introduction}

Modern robotic tasks in industrial settings demand highly accurate segmentation masks to facilitate robust grasp planning. 
However, traditional closed-set supervised models, such as YOLO~\cite{redmon2016yolo} or Mask R-CNN~\cite{he2017mask} are often unsuitable for these applications due to extreme data scarcity and the requirement for zero-shot generalization.
Annotating thousands of domain-specific images, such as specific metal alloys in recycling or rare plant phenotypes in precision farming, is economically impracticable, and in cases where industrial objects are unknown beforehand even impossible.
In dynamic environments, robotic systems frequently encounter objects that were not represented in the training distribution.

To address these limitations, recent advances in Vision-Language Models (VLMs) and self-supervised Vision Transformers (ViT)~\cite{dosovitskiy2020image} provide a promising alternative. 
By leveraging models such as DINOv3~\cite{simeoni2025dinov3} for dense feature extraction and SAM 3~\cite{carion2025sam3segmentconcepts} for geometry-aware segmentation, it is possible to identify anomalous clusters that deviate from a nominal background and object distribution.

\begin{figure}
    \centering
    \includegraphics[width=1\linewidth]{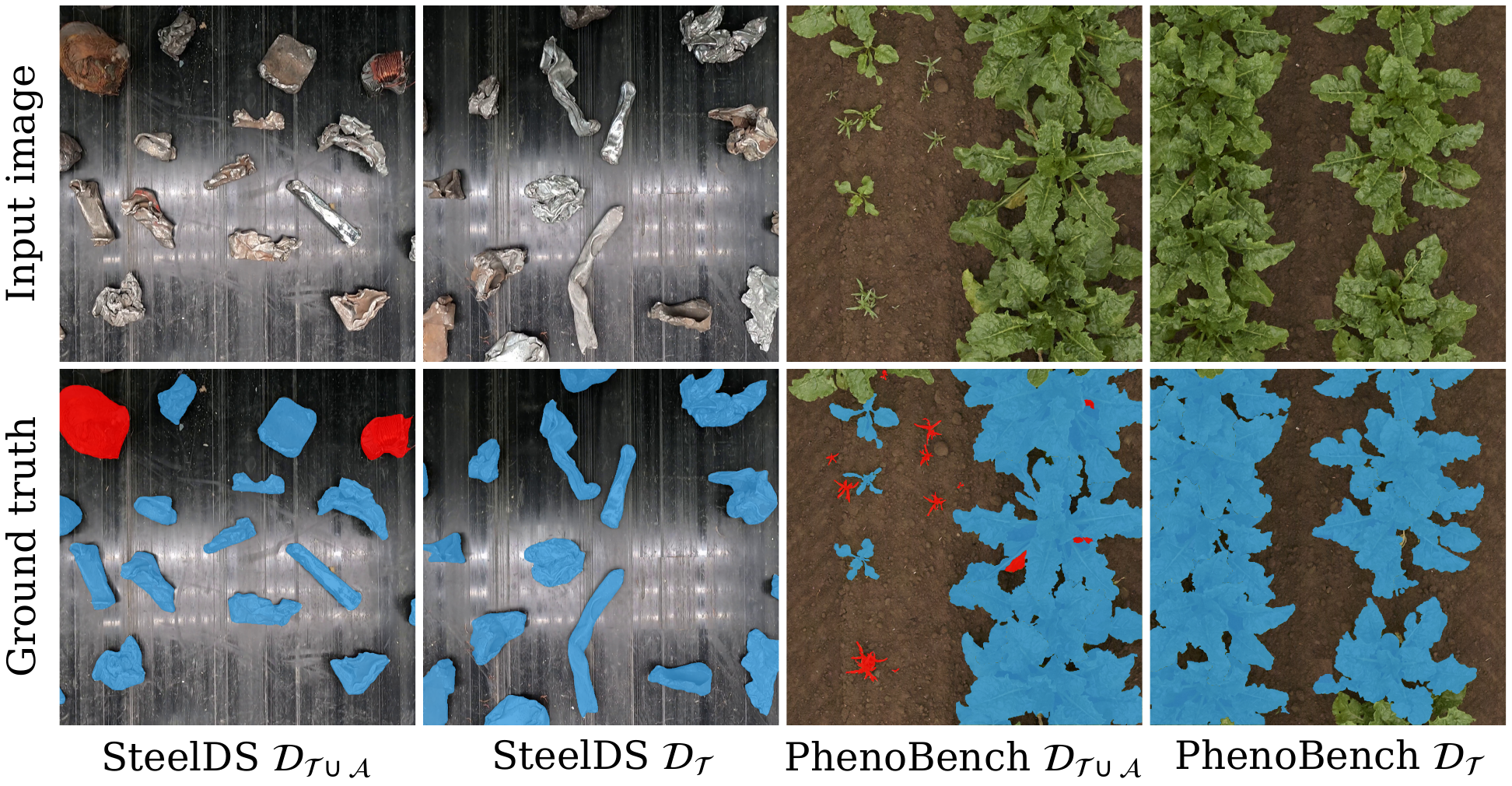}
    \caption{Dataset comparison for SteelDS and PhenoBench. Input images and ground truth segmentations (blue: \textit{target}, red: \textit{anomaly}) highlight the structural difference between the mixed baseline ($D_{\mathcal{T} \cup \mathcal{A}}$) and Anomaly-free reference ($D_{\mathcal{T}}$).}
    \label{fig:ds_example}
\end{figure}

In industrial robotic sorting, the primary objective is the identification and subsequent removal of specific items from a heterogeneous flow of materials. 
This task can be formally characterized as a change or anomaly detection and semantic segmentation problem between two distinct sets. 
The baseline dataset, contains a stochastic mixture of all object classes, including background, target items and anomalies (Fig. \ref{fig:ds_example}).
Conversely, the target dataset represents the same domain post-sorting, where specific classes are absent. 
The fundamental challenge lies in the absence of explicit supervision; the system lacks any manual annotations, such as pixel-level masks, bounding boxes, or even explicit image-level class labels. 
Instead, the only available information is the structural difference between the two datasets, necessitating a methodology capable of capturing semantic shifts in a weakly supervised manner.

This paper proposes a weakly supervised pipeline that improves zero-shot object localization through the following contributions:
(i) We replace domain-specific, hand-crafted heuristics in baseline approaches with open-vocabulary semantic text-prompting via SAM 3. 
This generalization eliminates the reliance on rigid geometric priors and enables robust applicability across diverse, unstructured environments.
(ii) Target and Anomaly objects are identified in a completely label-free manner by analyzing mode discrepancies (Fig.~\ref{fig:header}, middle) — specifically, isolating clusters that manifest in the observed scene but are missing in the nominal reference, within a quantized ViT feature space.

\section{Related Work}

\textit{Self-Supervised Representation Learning for Robotics:}
Traditional robotic perception relies heavily on supervised learning, requiring extensive domain-specific annotations. 
Recently, self-supervised Vision-Language Models (VLMs) such as CLIP \cite{radford2021learning} have enabled open-vocabulary image classification. 
However, contrastive VLMs inherently optimize for global image representations, leading to suboptimal performance in dense prediction tasks required for robotic grasping. 
Conversely, self-distillation architectures such as DINO \cite{caron2021emerging} and DINOv3 \cite{simeoni2025dinov3} produce high-resolution, semantically consistent feature maps. 
These dense embeddings exhibit emergent property localization without explicit supervision, making them highly suitable for extracting distinguishing features of unknown objects in unstructured environments.

\textit{Zero-Shot and Open-Vocabulary Segmentation:}
The Segment Anything Model (SAM) \cite{kirillov2023segment} and its successors, such as SAM 3 \cite{carion2025sam3segmentconcepts}, introduced robust zero-shot generalization for class-agnostic object segmentation. 
While highly effective at extracting geometric boundaries, these models lack intrinsic semantic understanding.
To address this, approaches such as Grounding DINO \cite{liu2024grounding} and SAM 3 couple spatial priors with VLMs to allow text-prompted segmentation.
Recent works increasingly adapt Vision-Language Models for training-free zero-shot anomaly detection \cite{zhang2025trainingfree}, for instance by enhancing CLIP's structural anomaly awareness \cite{ma2025aaclip} or utilizing it for coarse cross-modal localization before SAM mask refinement \cite{li2025clipsam}. 
In dynamic robotic applications, the exact class or physical description of an Anomaly (e.g., novel or foreign objects) is often unknown a priori.
Methods relying on explicit zero-shot text prompting struggle when anomalies fall entirely out-of-distribution or lack semantic definitions.

Our pipeline bypasses this limitation by using SAM 3 purely for objectness abstraction and relying on feature distribution analysis rather than specific semantic vocabulary to classify unknown instances.

\textit{Unsupervised Anomaly Detection:}
Standard unsupervised anomaly detection models, such as PatchCore \cite{roth2022towards} and PaDiM \cite{defard2021padim}, are widely used in industrial inspection benchmarks. 
These methods construct a normative feature space utilizing pre-trained CNNs or ViTs. 
However, they operate under the strict assumption that the initial training dataset is completely free of anomalies. 
This assumption fundamentally fails in unstructured environments where target and anomaly instances naturally co-occur (mixed baseline data)~\cite{tong2025comgen}.
Furthermore, identifying mere pixel-level anomalies is often insufficient; reliable robotic tasks require distinct anomaly instance segmentation to differentiate multiple unknown objects in complex scenes \cite{nekrasov2025oodis}.

In contrast, our weakly supervised approach leverages a comparative distribution analysis between a mixed baseline and a sparse nominal reference, thus eliminating both the need for purely nominal training data and the reliance on predefined textual anomaly prompts.

\textit{Robotic Applications and Cross-Domain Inspection:}
Automating material recovery through robotics is highly desirable not only to enhance industrial tasks efficiency but also to mitigate significant health risks for workers operating in highly contaminated, unstructured waste environments \cite{thilakarathna2025robotic}.
Currently, robotic tasks in agriculture and recycling heavily depends on domain-specific heuristics. 
For instance, agricultural weed removal systems frequently utilize vegetation indices such as Excess Green (ExG) \cite{woebbecke1995color} coupled with morphological operations to isolate crops from background soil. 
Recent weakly-supervised pipelines, such as WaW \cite{waw_phenobench}, leverage these heuristics to generate point prompts for SAM. 
While computationally efficient, these hand-crafted priors fail to generalize to industrial domains lacking specific chromatic markers, such as metallic scrap datasets~\cite{neubauer2025ibis}. 

By replacing rigid heuristics with generalized semantic prompting and ViT-based clustering, our method provides a unified, domain-agnostic inspection framework.


\section{Method}

\subsection{Problem Statement}

We address the problem of weakly supervised object discovery within a two-stage separation process. 
Let $\mathcal{D}_{\mathcal{T} \cup \mathcal{A}} = \{I_1, \dots, I_N\}$ denote the baseline dataset ('before separation'), where each image $I \in \mathbb{R}^{H \times W \times 3}$ contains a random variation of 
Target objects $\mathcal{T}$ and Anomaly objects $\mathcal{A}$.
Let $\mathcal{D}_{\mathcal{T}} = \{J_1, \dots, J_M\}$ denote the reference dataset ('after separation'), where each image $J \in \mathbb{R}^{H \times W \times 3}$ contains only the target elements from the same domain distribution $\mathcal{T}$.

Given weak supervision without pixel- or explicit image-level labels, let $\mathbf{f} \in \mathbb{R}^d$ denote a patch embedding extracted via a pre-trained Vision Transformer. 
We assume the reference dataset approximates the target feature distribution:

\begin{equation}
    P(\mathbf{f} | \mathcal{D}_{\mathcal{T}}) \approx P(\mathbf{f} | \mathcal{T})
\end{equation}
and that the feature distribution of the mixed baseline dataset is a linear combination of target and anomaly features:
\begin{equation}
    P(\mathbf{f} | \mathcal{D}_{\mathcal{T} \cup \mathcal{A}}) = (1 - \lambda)P(\mathbf{f} | \mathcal{T}) + \lambda P(\mathbf{f} | \mathcal{A})
\end{equation}
where $\lambda$ represents the unknown mixing ratio of anomaly objects in the baseline data.
The objective of PASTA is to extract a discrete segmentation mask $M \in \{0, 1, 2\}^{H \times W}$ for a given input image $I$.
A pixel $p$ is assigned to one of three categories based on its feature representation:

\begin{itemize}
    \item \textbf{Class 2 (Anomaly $\mathcal{A}$):} Assigned if the feature resides in a region of the latent space significantly populated in the mixed dataset $\mathcal{D}_{\mathcal{T} \cup \mathcal{A}}$ but sparse or empty in the reference dataset $\mathcal{D}_{\mathcal{T}}$.
    \item \textbf{Class 1 (Target $\mathcal{T}$):} Assigned to object features (e.g., crops, steel) that are semantically distinct and consistently present in both datasets.
    \item \textbf{Class 0 (Background):} Assigned to the rest of the image; it may contain diverse visual clutter (e.g., conveyor belts, soil, machine parts).
\end{itemize}

\subsection{Baseline Implementation}
\label{sec:baseline}
To establish a comparative baseline, we reconstructed and extend the anomaly detection pipeline proposed in recent related work~\cite{waw_phenobench} (WaW) for crop vs. weed segmentation.
This method frames the task as an unsupervised anomaly segmentation problem by curating a normative bag-of-features representing the prevalent crop plants. Utilizing the Segment Anything Model (SAM) and BioCLIP, representative plant features are extracted and aggregated via popularity voting. During inference, vegetation segments exhibiting low similarity to this established crop manifold are subsequently classified as anomalous weeds.
The fundamental assumption of this approach is that target objects exhibit distinct color profiles, specifically a high intensity in the green color channel, which is typical for agricultural domains.

\textbf{Heuristic Object Extraction:}
The baseline method relies on the Excess Green (ExG)~\cite{woebbecke1995color} vegetation index to isolate potential objects from the background. 
These isolated regions are utilized to generate point prompts for the Segment Anything Model (SAM) \cite{kirillov2023segment}, yielding instance masks for the detected objects. 
This heuristic is strictly domain-dependent. 
Consequently, the ExG approach fails to generalize to non-agricultural domains lacking distinct color markers, such as industrial tasks.

\textbf{Cross-Domain Generalization via SAM 3:}
To achieve cross-domain generalization, we replace the ExG heuristic with the Segment Anything Model 3 (SAM 3)~\cite{carion2025sam3segmentconcepts} guided by broad semantic text prompts. 
While SAM 3 may lack the specific vocabulary to classify unique materials or novel plant phenotypes zero-shot, it provides robust objectness priors. 
It reliably distinguishes physical entities from unstructured background clutter. 
By applying a domain-specific but class-agnostic text prompt (e.g., \textit{plants} for PhenoBench or \textit{objects on the conveyor belt} for SteelDS), SAM 3 generates a set of disjoint object masks $S = \{m_1, m_2, \dots, m_k\}$ for each image. 
These masks act as precise spatial boundaries, isolating relevant foreground objects independently of their specific color or material properties.

\begin{figure}[t]
    \centering
    \includegraphics[width=1\linewidth]{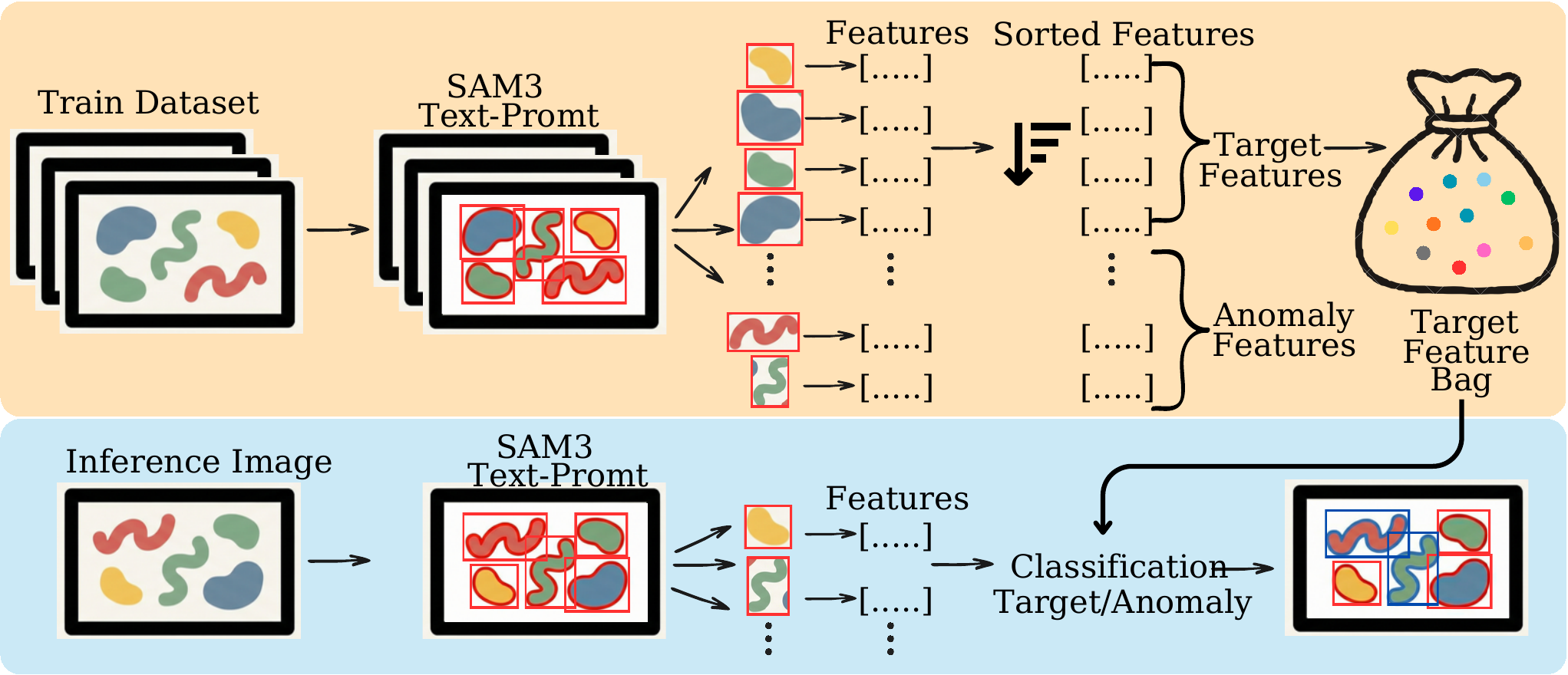}
    \caption{Reconstructed train and inference pipeline. 
    Training (top): SAM 3 extracts object patches, which are embedded and sorted by frequency.
    High-frequency embeddings define the Target Feature Bag, whereas low-frequency embeddings are isolated as anomalies. 
    Inference (bottom): Extracted features are classified against the target bag.
    }
    \label{fig:pipeline_waw_rebuild}
\end{figure}

\textbf{Feature Extraction and Anomaly Classification:}
Figure~\ref{fig:pipeline_waw_rebuild} illustrates the architecture of the reconstructed pipeline.
In this context, a feature is defined as a high-dimensional semantic embedding that encodes the visual properties of a specific object.
To obtain these representations, the input image is first filtered using the generated object masks, isolating the target regions from the background.
These isolated patches are subsequently processed by a pre-trained vision foundation model, which maps the pixel data into a dense latent space.
For the backbone, we evaluated CLIP and BioCLIP as proposed in the original study~\cite{waw_phenobench}, alongside with variations of the DINOv3 model.
The extracted feature vectors are then aggregated and sorted according to their frequency of occurrence across the dataset.
Anomaly classification relies on the assumption that rare features, those manifesting with a significantly lower frequency, correspond to the unseen anomaly classes.

\subsection{PASTA: Patch Aggregation for Segmentation of Targets and Anomalies}
\label{sec:our_method}
The existing unsupervised anomaly detection baseline~\cite{waw_phenobench} relies on domain-specific, hand-crafted heuristics, such as dedicated color indices or rigid spatial priors, which strictly limit their generalization to novel, unstructured environments.
To eliminate this dependency, we propose an pipeline for detecting anomalies in agricultural and industrial images by leveraging self-supervised Vision Transformers (ViT) and clustering-based distribution analysis.
The pipeline operates under the assumption that anomalies (e.g., weeds in a crop field or foreign objects on a conveyor belt) manifest as distinctive clusters that are present in a mixed baseline dataset $\mathcal{D}_{\mathcal{T} \cup \mathcal{A}}$ but absent in a clean reference dataset $\mathcal{D}_{\mathcal{T}}$.
This assumption is agnostic to the specific domain, while we will demonstrate on two distinct datasets: \textbf{PhenoBench}~\cite{phenobench_ds} (agricultural crop/weed segmentation) and \textbf{SteelDS} (industrial steel/copper segmentation~\cite{neubauer2025ibis}) (see Fig.~\ref{fig:ds_example}).

\begin{figure}[h]
    \centering
    \includegraphics[width=1\linewidth]{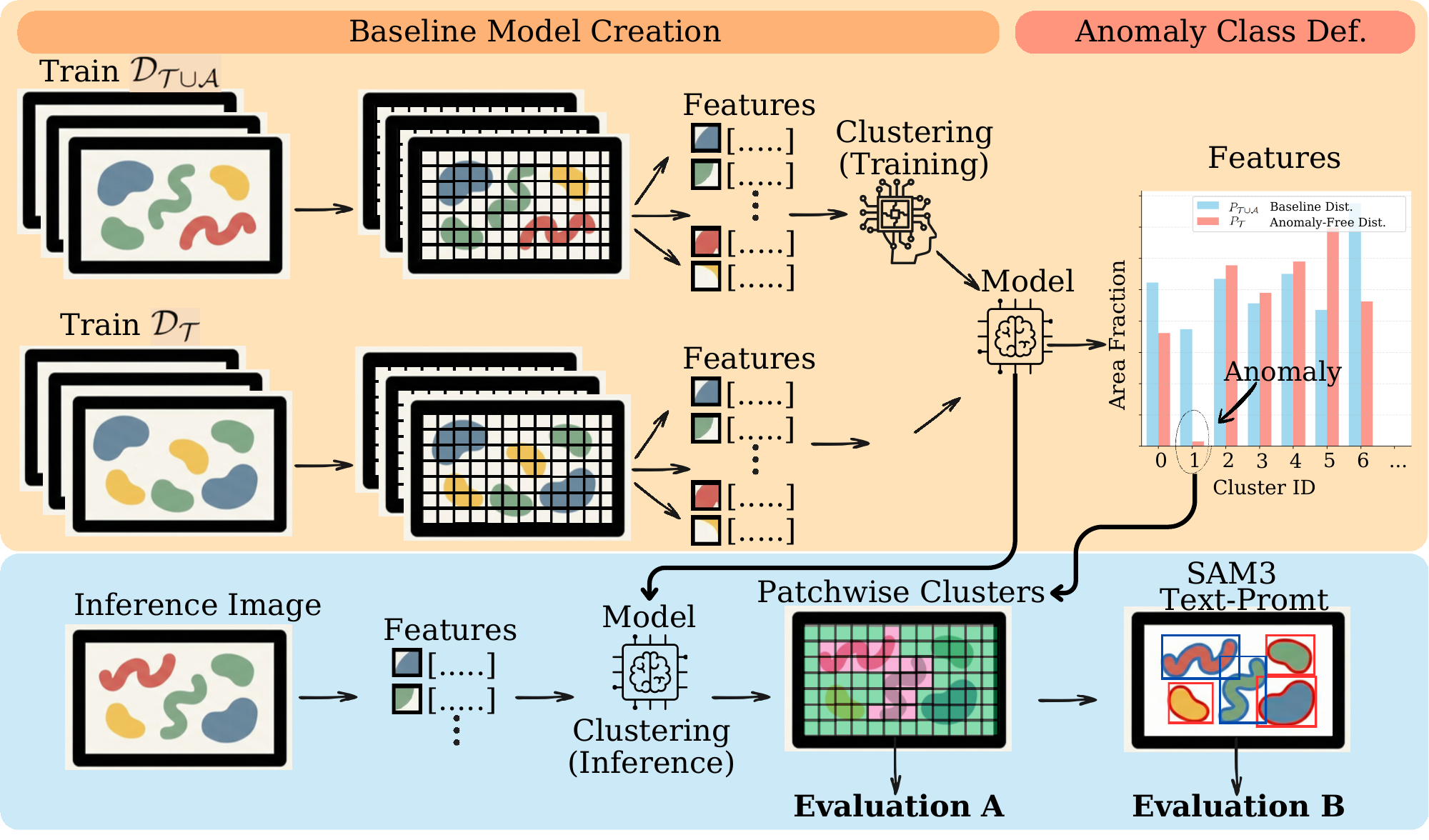}
    \caption{Architecture of PASTA. 
    The training phase (top) identifies anomaly-specific feature clusters via distribution analysis between a mixed baseline and an anomaly-free reference dataset. 
    The inference phase (bottom) applies the trained model for dense patchwise anomaly classification (Evaluation A in Section~\ref{evaluationA}), which is subsequently refined by SAM 3 semantic prompting for precise object segmentation (Evaluation B in Section~\ref{evaluationB}).}
    \label{fig:own_pipeline}
\end{figure}

The overall architecture of PASTA is illustrated in Figure~\ref{fig:own_pipeline}. 
    The model training consists of two primary phases: (1) \textit{Baseline Model Creation}, where a clustering model is fitted to the extracted features of the mixed baseline dataset $\mathcal{D}_{\mathcal{T} \cup \mathcal{A}}$, and (2) \textit{Anomaly Class Definition}, where anomalous clusters are identified by analyzing distributions in the clean reference dataset $\mathcal{D}_{\mathcal{T}}$.

\subsubsection{Phase 1: Baseline Model Creation (Weakly Supervised Feature Learning)}
The first phase aims to establish a baseline representation of the visual data by utilizing a pre-trained Vision Transformer (e.g., DINOv3, BioCLIP, CLIP) as a feature extractor.
Given a baseline dataset $\mathcal{D}_{\mathcal{T} \cup \mathcal{A}}$, the processing pipeline consists of three sequential steps.

\textbf{Feature Extraction:} Input images are resized and normalized before being passed through the ViT backbone. 
We extract patch-level embeddings from the last hidden layer of the transformer, resulting in a set of feature vectors $F_{\mathcal{T} \cup \mathcal{A}} = \{f_1, f_2, \dots, f_N\}$, where $f_i \in \mathbb{R}^d$ and $N$ is the total number of patches across all images in $\mathcal{D}_{\mathcal{T} \cup \mathcal{A}}$. 
The embedding dimension $d$ is determined by the specific Vision Transformer architecture utilized.

\textbf{Clustering:} To quantize this continuous feature space into interpretable semantic distinct units, we employ Mini-Batch K-Means clustering on $F_{\mathcal{T} \cup \mathcal{A}}$. The algorithm groups similar visual patterns (e.g., soil textures, crop leaves) into $K$ clusters.
\textbf{Reference Distribution:} Finally, we compute the probability distribution $P_{\mathcal{T} \cup \mathcal{A}}(K)$ of these clusters over the entire dataset $\mathcal{D}_{\mathcal{T} \cup \mathcal{A}}$, representing the expected frequency of each visual pattern. 
The trained K-Means model and $P_{\mathcal{T} \cup \mathcal{A}}$ constitute our baseline model.

\subsubsection{Phase 2: Anomaly Class Definition (Distribution Analysis)}
\label{phase2}
The second phase identifies which of the learned clusters correspond to the ``Target'' (normal) class and which correspond to ``Anomalies''. 
We introduce a second, anomaly-free dataset, (e.g., a ``weed-free'' crop field or a ``copper-free'' conveyor belt).

\textbf{Target Distribution Estimation:} 
We apply the frozen feature extractor and K-Means model from Phase 1 to $D_{\mathcal{T}}$, projecting its content into the same cluster space to compute the target distribution $P_{\mathcal{T}}(K)$ (Figure~\ref{fig:header}, orange histogram bars).
\textbf{Anomaly Identification via Missing Features:} Anomalies are identified based on the principle of missing features. 
Clusters prominent in the mixed baseline set $P_{\mathcal{T} \cup \mathcal{A}}$ (see Figure~\ref{fig:header} the blue bars) but significantly suppressed in the clean target set $P_{\mathcal{T}}$ are flagged. 
\textbf{Ratio Calculation:} For each cluster $k_i$, we calculate the ratio:
\begin{equation}
    R_i = \frac{P_{\mathcal{T}}(k_i)}{P_{\mathcal{T} \cup \mathcal{A}}(k_i)}
\end{equation}
Clusters with $R_i \approx 0$ (or below a defined threshold) imply that the visual pattern is present in the $D_{\mathcal{T} \cup \mathcal{A}}$ but absent in $D_{\mathcal{T}}$ (see Fig.~\ref{fig:ds_example}). 
These identified clusters are stored as the anomaly clusters, while all other clusters are considered normal target clusters (e.g., crop, soil).

\subsubsection{Phase 3: SAM 3-Enhanced Object Classification (Mask-Feature Fusion)}
\label{phase3}
While the patch-level analysis in Phase 2 only classifies anomalies at the coarse spatial resolution of the ViT feature grid, industrial and agricultural applications typically require pixel-aligned, object-level instance segmentation.
To bridge this gap, we integrate the Segment Anything Model 3 (SAM 3)~\cite{carion2025sam3segmentconcepts} to extract precise semantic boundaries and classify these instances based on the underlying patch clusters.

We prompt SAM 3 with a generic, domain-specific text prompt (e.g., \textit{plants} or \textit{objects on the conveyor belt}) to generate a set of disjoint instance masks $S = \{m_1, m_2, \dots, m_k\}$, where $m_j \in \{0, 1\}^{H \times W}$, for all potential objects in a given image.
Simultaneously, the ViT pipeline provides a coarse cluster map containing the assigned cluster IDs per image patch.
For each SAM3 generated mask $m_j$, we overlay its spatial boundaries onto the cluster map to analyze the underlying cluster distribution.
We compute the histogram of cluster IDs located within the foreground area of $m_j$ and calculate the proportion of the area assigned to anomaly clusters.
If this anomaly ratio exceeds a predefined threshold $\gamma$, the entire segment $m_j$ is classified as an anomaly in the final output mask.
Otherwise, the entire segment is uniformly classified as a normal target object.

\subsection{Implementation Details}
All experiments were executed on an NVIDIA RTX 4090 GPU with 24GB VRAM running Ubuntu 24.04 and averaged over 5 seeds to account for stochastic variations during clustering. 
Input images were preprocessed using standard PyTorch transformations, which included resizing, conversion to tensor format, and normalization.
Image resolutions were adapted dataset-specifically. 
For PhenoBench, images were resized to $1024 \times 1024$ pixels. 
For SteelDS, original high-resolution images ($3840 \times 2160$ pixels) were resized to $910 \times 512$. 
The patch size for feature extraction was dynamically determined by the respective Vision Transformer (ViT) backbone architecture utilized in all the experiments. 
In Phase 2~\ref{phase2}, the threshold for anomaly identification was empirically set to $R_i <$ 0.05. For the SAM 3 enhanced classification (Phase 3~\ref{phase3}), the mask-feature fusion voting threshold was defined as $\gamma = $ 0.1 (= 10\%).

\section{Evaluation}

To assess the performance of PASTA, our experimental setup is structured into three distinct evaluations:
first, a comparison against a heuristic baseline; second, an assessment of the isolated patch-level classification (Evaluation A); and third, an analysis of the segment-based refinement (Evaluation B).  
All evaluations utilize manually annotated ground truth masks where pixels are labeled as Class 1 (Target), Class 2 (Anomaly), or Class 0 (Background). 
Segmentation performance is quantified using Intersection-over-Union (IoU) in [\%].

The evaluations are conducted on three distinct datasets, i.e., PhenoBench (agricultural crop/weed), SteelDS (industrial steel/copper), and SteelDS Extended (industrial dataset, more and smaller objects than SteelDS), listed in Table~\ref{tab:datasets}.

\begin{table}[h]
    \centering
    \caption{Number of samples used for training ($\mathcal{D}_{\mathcal{T} \cup \mathcal{A}}$, $\mathcal{D}_{\mathcal{T}}$) and testing ($\mathcal{D}_{test}$) on the two datasets.}
    \label{tab:datasets}
    \begin{tabular}{lrrr}
        \toprule
        \textbf{Dataset} & $|\mathcal{D}_{\mathcal{T} \cup \mathcal{A}}|$ & $|\mathcal{D}_{\mathcal{T}}|$ & $|\mathcal{D}_{test}|$ \\
        \midrule
        PhenoBench & 1407 & 1407 & 772 \\
        SteelDS & 3831 & 900 & 956 \\
        SteelDS Extended & 16854 & 3677 & 3742 \\
        \bottomrule
    \end{tabular}
\end{table}

\begin{table}[t]
    \centering
    \caption{Average results (mIoU for \textbf{B}ackground, \textbf{T}arget, \textbf{A}nomaly) of the reconstructed baseline and its variations (\textbf{P}heno\textbf{B}ench, \textbf{S}teel\textbf{DS}) over five random seeds. }
    \label{tab:av_exp1_miou}
    \setlength{\tabcolsep}{3pt}
    \begin{tabular}{l l l r r c c c c c c}
        \toprule
        \textbf{Approach} & \textbf{Backbone} & \textbf{IoU B} & \textbf{IoU T} & \textbf{IoU A} & \textbf{mIoU} \\
        \midrule
        PB (WaW)& BioCLIP & 98.85 & 68.06 & \textcolor{red}{11.13} & 59.34 \\
        \midrule
        \multirow{5}{*}{SDS (Ours)} & BioCLIP & 99.35 & 64.20 & 35.67 & 66.41 \\
        & CLIP ViT-B-16 & 99.35 & 74.81 & 8.80 & 60.98 \\
        & CLIP ViT-B-32 LAION2b & 99.35 & \textbf{86.20} & \textbf{54.34} & \textbf{79.96} \\
        & DINOv3 ConvNeXt-Tiny & 99.35 & 44.86 & 27.84 & 57.35 \\
        & DINOv3 ViT-Small & 99.35 & 39.93 & 24.06 & 54.44 \\
        \midrule
        \multirow{5}{*}{PB (Ours)} & BioCLIP & 98.59 & 44.93 & 6.76 & 50.09 \\
        & CLIP ViT-B-16 & 98.59 & 48.72 & 5.23 & 50.85 \\
        & CLIP ViT-B-32 LAION2b & 98.59 & \textbf{51.83} & 6.89 & \textbf{52.43} \\
        & DINOv3 ConvNeXt-Tiny & 98.59 & 35.59 & \textbf{7.61} & 47.27 \\
        & DINOv3 ViT-Small & 98.59 & 38.71 & 5.97 & 47.76 \\
        \bottomrule
    \end{tabular}
\end{table}

\begin{table}[t]
    \centering
    \caption{Feature patch extraction methods for PhenoBench and SteelDS. Our approach produce fewer but more relevant patches, resulting in a reduction of the training time by more than 70\%.}
    \label{tab:av_exp1_time}
    \setlength{\tabcolsep}{3pt}
    \begin{tabular}{l l l r r c c c c c c}
        \toprule
        \textbf{Dataset (Pipeline)} & \textbf{Seg. Model} & \textbf{Patches} & $k_{sphere}$ & $k_v$ & \textbf{Time} [h] \\
        \midrule
        PhenoBench (WaW~\cite{waw_phenobench})& SAM & 64330 & 100 & 10 & $\approx$ 2.8 \\
        SteelDS (Ours) & SAM 3 & 11247 & 260 & 10 & \textbf{$\approx$ 0.8} \\
        PhenoBench (Ours) & SAM 3 & 15565 & 120 & 10 & \textbf{$\approx$ 0.7} \\
        \bottomrule
    \end{tabular}
\end{table}

\subsection{Baseline Evaluation}
\label{sec:baseline_eval}
We start by evaluating the proposed SAM~3-enhanced anomaly detection pipeline against the heuristic-based baseline. 
We utilize the state-of-the-art zero-shot method, Weeds Are Weird (WaW)~\cite{waw_phenobench}, as the primary reference to assess computational efficiency, zero-shot generalization, and parameter robustness in robotic inspection tasks.
Note that all PhenoBench results are derived from the number of validation samples. 
The official test set is not publicly available.

\textbf{Computational Efficiency and Feature Abstraction:}
A fundamental architectural shift in our approach is the replacement of the ExG heuristic with open-vocabulary semantic text-prompting via SAM~3 (see Section~\ref{sec:baseline}). 
While WaW relies on dense point prompts derived from ExG, which frequently results in over-segmentation, SAM~3 isolates complete semantic object instances.
This spatial abstraction drastically reduces the volume of extracted feature embeddings. 
As detailed in Table~\ref{tab:av_exp1_time}, the number of extracted feature patches on the PhenoBench dataset drops by 75.8\% (from 64,330 to 15,565). 
This reduction translates directly to improved computational efficiency, decreasing the total processing time from approximately 2.8~h to 0.7~h, a critical requirement for real-time robotic applications.

\textbf{Zero-Shot Generalization Performance:}
We apply various pre-trained foundation models to unstructured industrial (SteelDS) and agricultural (PhenoBench) environments,  to evaluate their out-of-distribution (OOD) zero-shot capabilities.
On the SteelDS dataset, the \textit{CLIP LAION2b} backbone demonstrates robust generalization, segmenting 86.20\% of Target objects and detecting 54.34\% of Anomalies (see Table~\ref{tab:av_exp1_miou}).
Conversely, performance on PhenoBench is constrained.
The best-performing variation, \textit{DINOv3 ConvNeXt-Tiny}, achieves an Anomaly IoU of 7.61\%, underperforming the domain-specific WaW baseline (11.13\%).
This indicates that while holistic object-level features reduce computational overhead and enable cross-domain transfer without architectural adaptations, they sacrifice the fine-grained discriminative resolution required to distinguish subtle intra-domain variations (e.g., specific weed phenotypes vs. crops).

\begin{figure}[h]
    \centering
    \begin{subfigure}{\linewidth}
        \centering
        \includegraphics[width=\linewidth]{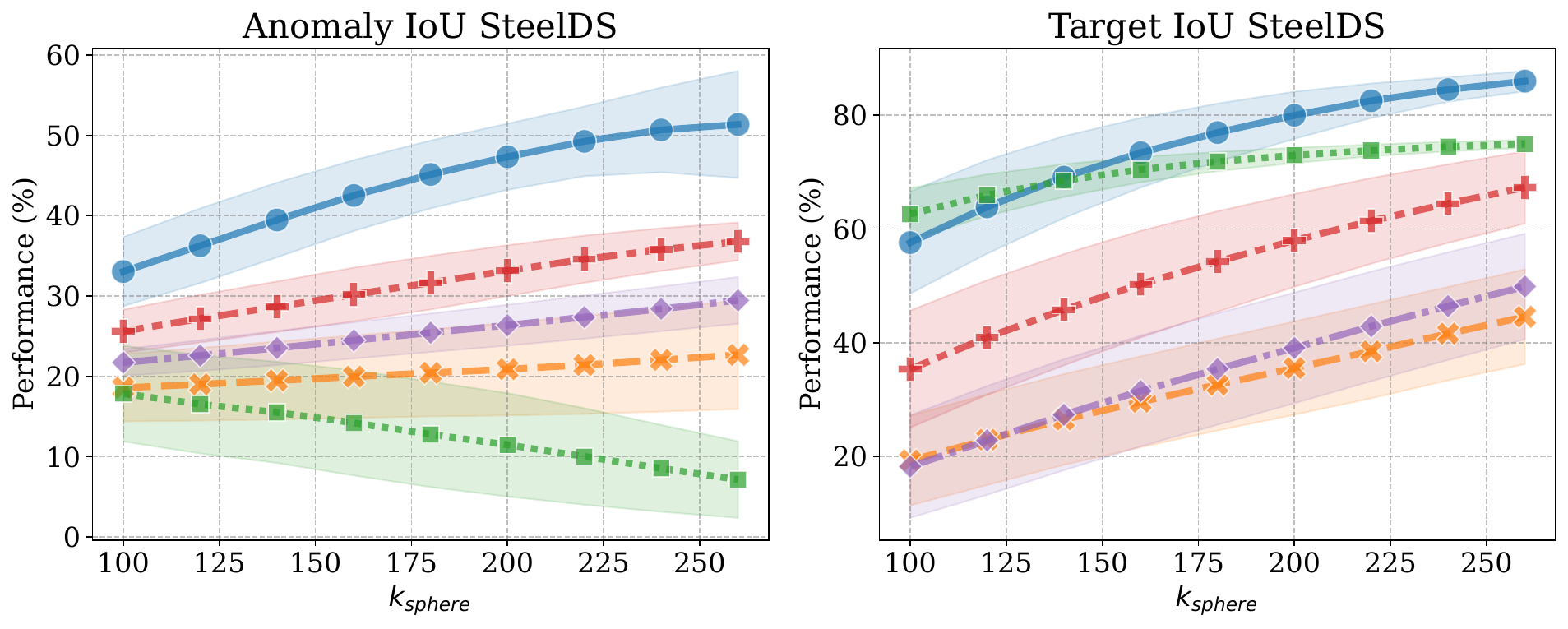}
    \end{subfigure}
    
    
    \begin{subfigure}{\linewidth}
        \centering
        \includegraphics[width=\linewidth]{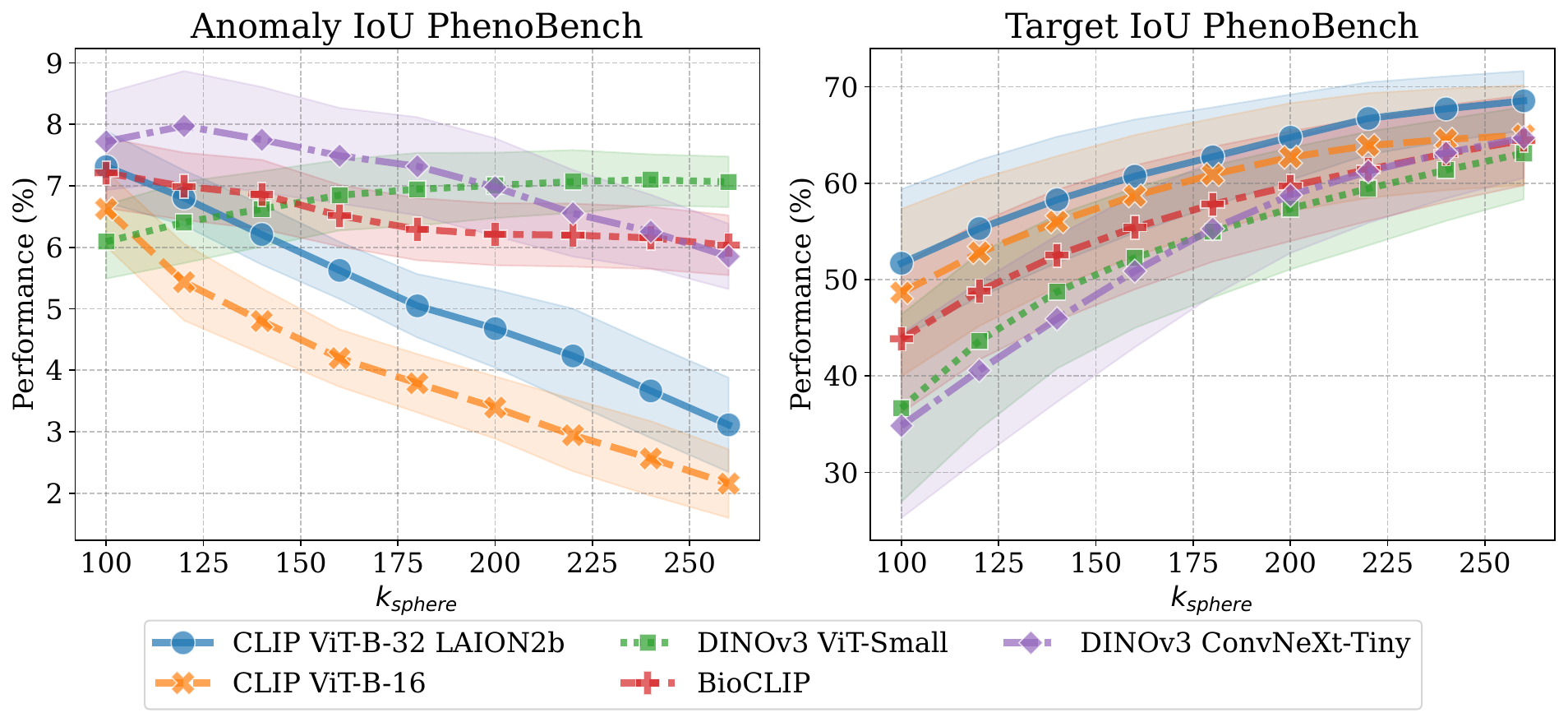}
    \end{subfigure}
    
    \caption{Backbone Robustness: Impact of hypersphere density ($k_{sphere}$) on Target and Anomaly IoU. 
    Shaded regions represent the standard deviation over 5 seeds. 
    In the industrial domain (SteelDS, top), both Target and Anomaly accuracy generally improve with larger $k_{sphere}$. 
    In the agricultural domain (PhenoBench, bottom), while Target accuracy generally improves, Anomaly accuracy often degrades as the expanding hypersphere starts encompassing anomaly features.}
    \label{fig:robustness_combined}
\end{figure}

\begin{figure}[h]
    \centering
    \begin{subfigure}{0.49\columnwidth}
        \centering
        \includegraphics[width=\textwidth]{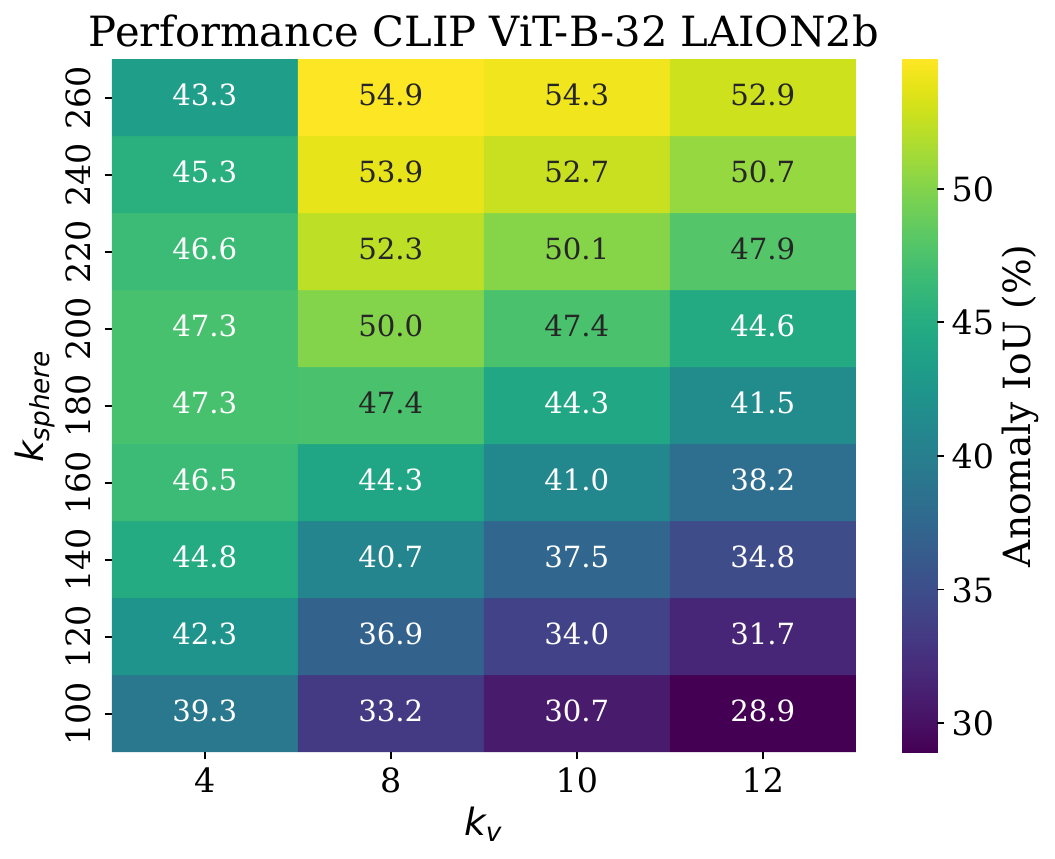}
        \captionsetup{justification=centering}
        \caption{SteelDS}
        \label{fig:heatmap_steel_sub}
    \end{subfigure}
    \begin{subfigure}{0.49\columnwidth}
        \centering
        \includegraphics[width=\textwidth]{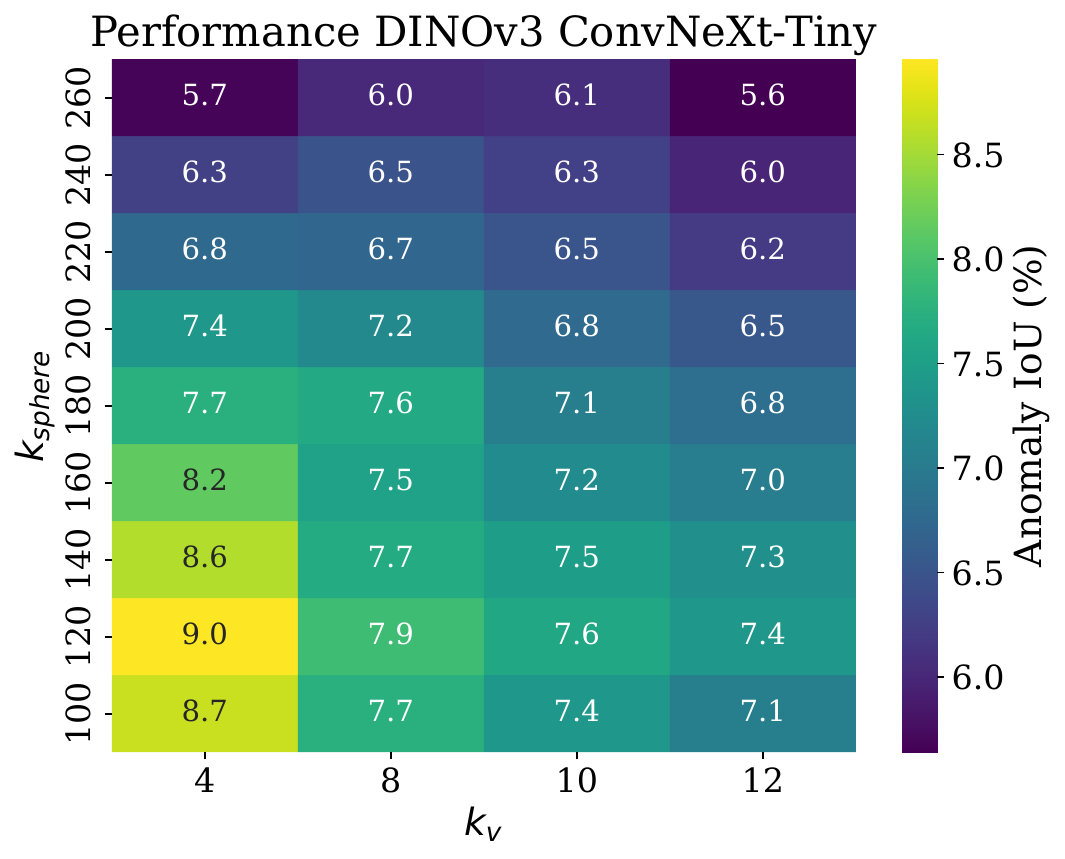}
        \captionsetup{justification=centering}
        \caption{PhenoBench}
        \label{fig:heatmap_pheno_sub}
    \end{subfigure}
    
    \caption{Parameter Sensitivity Heatmaps illustrating Anomaly IoU sensitivity to Density ($k_{sphere}$) versus Feature Voting K ($k_{v}$). (a) SteelDS benefits from high density, whereas (b) PhenoBench requires constrained density to maintain discriminative power against anomalies.}
    \label{fig:heatmaps_combined}
\end{figure}

\textbf{Robustness and Parameter Sensitivity:}
To analyze the stability of the ViT backbones, we evaluated the impact of two core hyperparameters adapted from the baseline methodology~\cite{waw_phenobench}: the hypersphere density parameter ($k_{sphere}$) and the Feature Voting K ($k_{v}$). 
$k_{sphere}$ specifies the radius of the nominal feature hyperspheres, calculated as the Euclidean distance to the $k$-th nearest neighbor in the reference feature space, which controls the inclusion boundary for the normal Target class. 
The parameter Feature Voting K ($k_v$) defines the neighborhood size required during the feature matching process to finalize a classification. 
We evaluated their combined impact on segmentation performance in Fig.~\ref{fig:robustness_combined} and \ref{fig:heatmaps_combined}.

\begin{itemize}
    \item \textbf{Target Accuracy:} Across both datasets and all backbones, Target IoU exhibits a strong positive correlation with $k_{sphere}$ (Fig.~\ref{fig:robustness_combined}). Expanding the hypersphere enables the model to capture greater intra-class variance of nominal objects, such as varying lighting conditions on metal or diverse crop geometries.
    \item \textbf{Anomaly Accuracy (SteelDS):} In domains with high visual inter-class variance, larger feature spaces are beneficial. As shown in Fig.~\ref{fig:robustness_combined} on the top, Anomaly IoU for CLIP LAION2b improves alongside Target IoU as $k_{sphere}$ increases. The parameter heatmap (Fig.~\ref{fig:heatmap_steel_sub}) confirms optimal performance at high density ($k_{sphere} \ge 240$) and moderate Feature Voting K ($k_v \le 10$).
    \item \textbf{Anomaly Accuracy (PhenoBench):} In domains with high visual similarity between target and anomaly, expanding the normal feature space is detrimental. Fig.~\ref{fig:robustness_combined} on the bottom demonstrates a monotonic decrease in Anomaly IoU as $k_{sphere}$ increases. The DINOv3 heatmap (Fig.~\ref{fig:heatmap_pheno_sub}) indicates that the model requires strictly constrained density values ($k_{sphere} \le 160$) to prevent the target hypersphere from absorbing visually similar anomaly features, thereby preserving discriminative power.
\end{itemize}

In conclusion, semantic object-level abstraction via SAM~3 provides substantial gains in cross-domain applicability and processing speed. 
However, for highly complex tasks with minimal inter-class variance, purely holistic feature extraction is insufficient, necessitating a hybrid approach that reincorporates local patch-level granularity.

\subsection{Patch-Based Classification (Evaluation A)}
\label{evaluationA}

To evaluate the isolated classification performance of the raw feature clusters without explicit object boundaries (Sec. \ref{sec:our_method}), we used patch-level features of test images which are mapped to their nearest cluster using the reference K-Means model. 
A critical limitation at this stage is the absence of object-level boundaries; the system can only broadly categorize patches as anomalies (Class 2) or nominal features. 
It is intrinsically impossible to differentiate between the specific target objects (Class 1) and the background (Class 0) based solely on patch clusters. 
To compute metrics against the ground truth, the resulting low-resolution patch map is upsampled to the original image resolution utilizing nearest-neighbor interpolation. 
Consequently, this evaluation focuses exclusively on the Anomaly IoU.

Table~\ref{tab:eval_a} shows a comparison of our patch-based approach for the two datasets. 
Performance contrasts sharply across models evaluated on the \textit{SteelDS} and between both datasets.
While CLIP almost completely fails to isolate anomalies at this coarse patch level in \textit{SteelDS}, yielding Anomaly IoU scores near zero, self-distilled DINOv3 architectures demonstrate superior dense feature representations. DINOv3 ViT-Small achieves here the highest performance (35.4\% Anomaly IoU with $K=20$), followed closely by DINOv3 ConvNeXt-Tiny (31.1\% Anomaly IoU with $K=15$). The high variance across random seeds, indicates that patch-level K-Means clustering is highly sensitive to initialization.
Anomaly detection in \textit{PhenoBench}, representative for the agricultural domain, presents a much harder task due to the extreme visual similarity between crops and weeds. Performance drops drastically across all models. On \textit{PhenoBench}, DINOv3 ConvNeXt-Tiny is the only backbone that manages to extract a marginal anomaly signal, reaching a peak Anomaly IoU of 8.6\% at $K=25$ clusters. The dense patch representations struggle to capture the fine-grained morphological differences required to separate anomalies in this domain.

The overall low performance, combined with the inability to separate target objects from the background, highlights the fundamental deficit of pure patch-based clustering. 
The coarse spatial resolution of the ViT feature grid intrinsically blends object boundaries and background clutter into mixed patches. 
This necessitates the integration of SAM 3 in the subsequent evaluation phase to extract precise spatial instances and resolve the ambiguity between target objects and the background.

\begin{table}[h]
\centering
\caption{Patch-level Anomaly IoU [\%] on SteelDS and PhenoBench across varying cluster counts ($K$). Results are averaged over 5 seeds, with the standard deviation provided in brackets. 
DINOv3 architectures significantly outperform contrastive models, peaking at 35.4\% IoU for SteelDS.}

\setlength{\tabcolsep}{4pt} 
\begin{tabular}{lcccc}
\toprule
 & \textbf{K=10} & \textbf{K=15} & \textbf{K=20} & \textbf{K=25} \\
\midrule

\multicolumn{5}{l}{\textbf{SteelDS}} \\
CLIP ViT-B-16
& 0.0 {\scriptsize (0.0)}
& 0.3 {\scriptsize (0.3)}
& 0.3 {\scriptsize (0.4)}
& 1.6 {\scriptsize (1.0)} \\

CLIP ViT-B-32 L.2b
& 2.7 {\scriptsize (1.5)}
& 2.4 {\scriptsize (2.1)}
& 2.6 {\scriptsize (1.4)}
& 2.7 {\scriptsize (1.5)} \\

CLIP ViT-B-32
& 0.0 {\scriptsize (0.0)}
& 0.2 {\scriptsize (0.3)}
& 0.3 {\scriptsize (0.5)}
& 0.4 {\scriptsize (0.6)} \\

DINOv3 ConvN.-T.
& 15.3 {\scriptsize (11.7)}
& \textbf{31.1 {\scriptsize (5.3)}}
& 25.1 {\scriptsize (11.7)}
& 23.8 {\scriptsize (13.5)} \\

DINOv3 ViT-Small 
& \textbf{20.7 {\scriptsize (15.8)}}
& 29.6 {\scriptsize (10.1)}
& \textbf{35.4 {\scriptsize (2.2)}}
& \textbf{34.0 {\scriptsize (2.0)}} \\

BioCLIP
& 0.0 {\scriptsize (0.0)}
& 0.6 {\scriptsize (1.4)}
& 0.0 {\scriptsize (0.0)}
& 0.0 {\scriptsize (0.0)} \\

\midrule
\multicolumn{5}{l}{\textbf{PhenoBench}} \\
CLIP ViT-B-16
& 0.7 {\scriptsize (0.1)}
& 0.6 {\scriptsize (0.1)}
& 0.6 {\scriptsize (0.2)}
& 0.6 {\scriptsize (0.1)} \\

CLIP ViT-B-32 L.2b
& 0.2 {\scriptsize (0.1)}
& 0.2 {\scriptsize (0.1)}
& 0.2 {\scriptsize (0.1)}
& 0.1 {\scriptsize (0.1)} \\

CLIP ViT-B-32
& 0.6 {\scriptsize (0.3)}
& 0.6 {\scriptsize (0.2)}
& 0.7 {\scriptsize (0.2)}
& 0.4 {\scriptsize (0.1)} \\

DINOv3 ConvN.-T.
& \textbf{4.2 {\scriptsize (2.4)}}
& \textbf{6.2 {\scriptsize (1.5)}}
& \textbf{7.6 {\scriptsize (0.7)}}
& \textbf{8.6 {\scriptsize (1.1)}} \\

DINOv3 ViT-Small 
& 2.9 {\scriptsize (2.1)}
& 2.6 {\scriptsize (2.2)}
& 4.0 {\scriptsize (2.8)}
& 5.1 {\scriptsize (1.8)} \\

BioCLIP
& 0.1 {\scriptsize (0.1)}
& 0.3 {\scriptsize (0.3)}
& 0.4 {\scriptsize (0.2)}
& 0.3 {\scriptsize (0.3)} \\

\bottomrule
\end{tabular}
\label{tab:eval_a}
\end{table}

\subsection{Segment-Based Classification (Evaluation B)}
\label{evaluationB}

\begin{figure}
    \centering
    \includegraphics[width=0.7\linewidth]{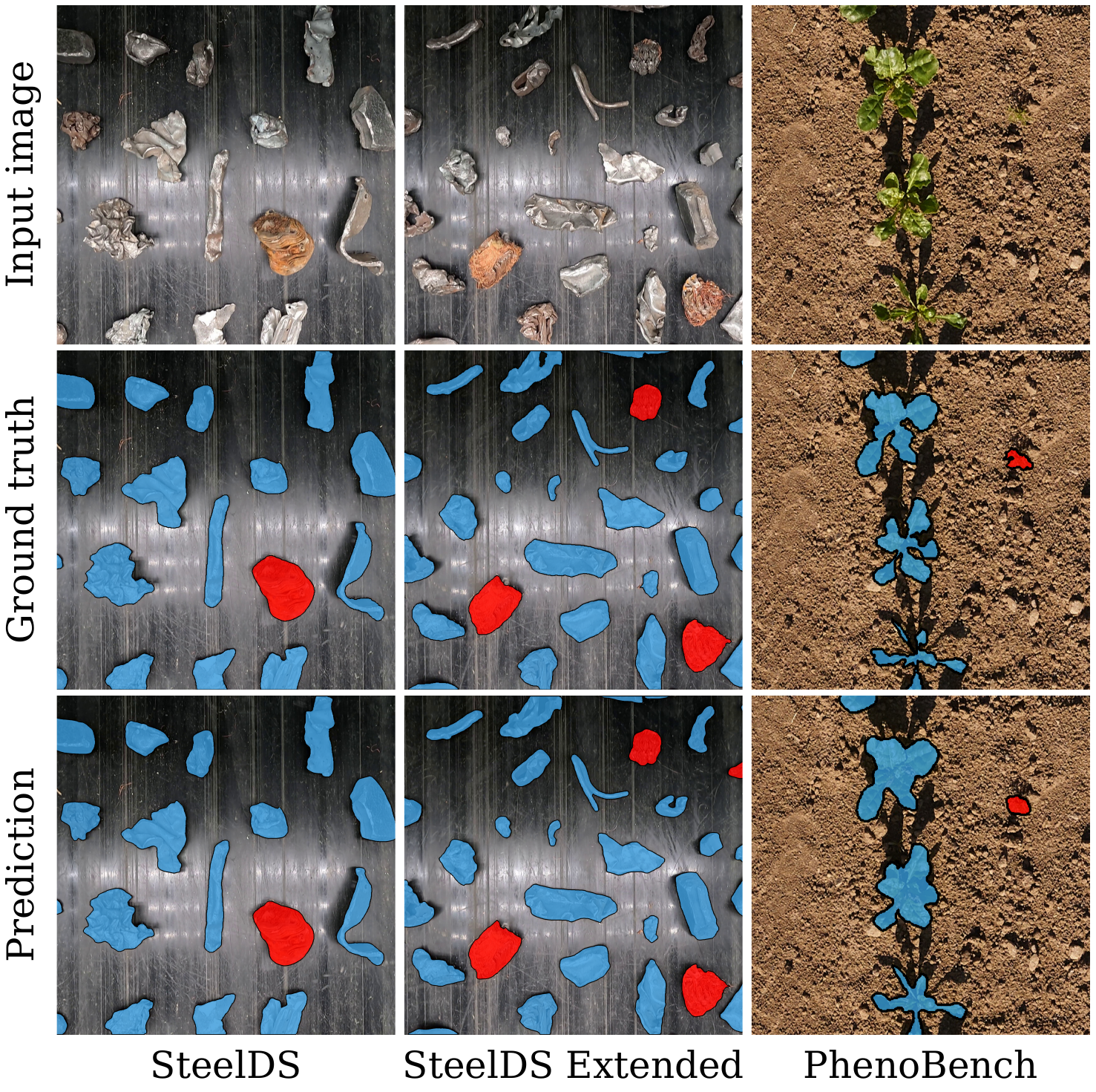}
    \caption{Qualitative segmentation results. Top: original image, middle: ground truth segments, bottom: PASTA predictions. blue: \textit{target}, red: \textit{anomaly}.}
    \label{fig:ds_example_final}
\end{figure}

Finally, we evaluate PASTA using the segment-based refinement (Sec.~\ref{sec:our_method}). 
Compared to the patch-level approach, this method aggregates patch-level votes within segments and results in more refined boundaries (Fig.~\ref{fig:ds_example_final}), as required for many robotics tasks. 
With this vote aggregation, we are able to segment both anomalies and target objects, and evaluate them based on entire object segments.

The IoU evaluation across datasets, VLMs, and cluster parameters in Table~\ref{tab:iou_evaluation}, compared to the patch-level results in Table~\ref{tab:eval_a}, shows that segment refinement provides a significant improvement in segmentation over raw and coarse patches. 
This improvement in segmentation performance is traded off against inference time for both evaluations (Table~\ref{tab:eval_time}).
Where the pure patch-based approach (Evaluation A) operates approximately three times faster than the segment-based classification (Evaluation B).

\begin{table}[t]
\centering
\caption{Training and inference time analysis. \textit{Model Setup} covers the training time, while \textit{Inference} is the processing time per image during test on the two approaches (\textbf{Ev}aluation \textbf{A} and \textbf{Ev}aluation \textbf{B}).}
\setlength{\tabcolsep}{4pt}
\begin{tabular}{lccc}
\toprule
 & \makecell{\textbf{Model} \\ \textbf{Setup [s]}} &\makecell{\textbf{Inference} \\ \textbf{Ev. A [ms]}} & \makecell{\textbf{Inference} \\ \textbf{Ev. B [ms]}} \\
\midrule

\multicolumn{4}{l}{\textbf{SteelDS}} \\
CLIP ViT-B-16
& 27.55 & 112 & 346 \\
CLIP ViT-B-32 L.2b
& \textbf{25.30} & \textbf{96} & \textbf{332} \\
CLIP ViT-B-32
& 27.08 & \textbf{96} & \textbf{333} \\
DINOv3 ConvN.-T.
& 29.57 & \textbf{97} & \textbf{332} \\
DINOv3 ViT-Small 
& 30.21 & 104 & 339 \\
BioCLIP
& 28.10 & 111 & 344 \\

\midrule
\multicolumn{4}{l}{\textbf{PhenoBench}} \\
CLIP ViT-B-16
& 6.68 & 41 & 207 \\
CLIP ViT-B-32 L.2b
& \textbf{5.63} & \textbf{30} & \textbf{197} \\
CLIP ViT-B-32
& 5.81 & \textbf{31} & \textbf{197} \\
DINOv3 ConvN.-T.
& 10.14 & 33 & 199 \\
DINOv3 ViT-Small 
& 10.42 & 36 & 203 \\
BioCLIP
& 7.19 & 42 & 207 \\

\bottomrule
\end{tabular}
\label{tab:eval_time}
\end{table}

\begin{table*}[t]
\centering
\caption{Hierarchical IoU evaluation for Target and Anomaly performance. All values are given in percent (\%), with standard deviations in brackets. Best and second best values per dataset and metric are marked in bold and underlined.}
\setlength{\tabcolsep}{4pt}
\footnotesize 
\begin{tabular}{l cccc cccc}
\toprule
& \multicolumn{4}{c}{\textbf{Target IoU}} & \multicolumn{4}{c}{\textbf{Anomaly IoU}} \\
\cmidrule(lr){2-5} \cmidrule(lr){6-9}
\textbf{Model} & \textbf{K=10} & \textbf{K=15} & \textbf{K=20} & \textbf{K=25} & \textbf{K=10} & \textbf{K=15} & \textbf{K=20} & \textbf{K=25} \\
\midrule

\multicolumn{9}{l}{\textbf{SteelDS}} \\
CLIP ViT-B-16
& 76.5 {\scriptsize (0.0)} & 65.3 {\scriptsize (13.0)} & 64.9 {\scriptsize (15.2)} & 31.7 {\scriptsize (17.8)} 
& 0.0 {\scriptsize (0.0)} & 1.7 {\scriptsize (2.4)} & 2.2 {\scriptsize (3.4)} & 12.6 {\scriptsize (4.6)} \\
CLIP ViT-B-32 L.2b
& 62.0 {\scriptsize (7.1)} & 62.9 {\scriptsize (14.5)} & 59.9 {\scriptsize (16.0)} & 58.9 {\scriptsize (13.3)} 
& 13.7 {\scriptsize (5.6)} & 11.1 {\scriptsize (9.2)} & 13.1 {\scriptsize (4.1)} & 13.2 {\scriptsize (4.7)} \\
CLIP ViT-B-32
& 76.4 {\scriptsize (0.1)} & 75.4 {\scriptsize (2.3)} & 75.9 {\scriptsize (0.9)} & 74.8 {\scriptsize (1.7)} 
& 0.1 {\scriptsize (0.2)} & 1.2 {\scriptsize (2.2)} & 1.5 {\scriptsize (2.7)} & 2.5 {\scriptsize (3.7)} \\
DINOv3 ConvN.-T.
& 46.6 {\scriptsize (28.9)} & 80.3 {\scriptsize (15.2)} & 83.6 {\scriptsize (8.4)} & 82.7 {\scriptsize (9.2)} 
& 23.2 {\scriptsize (17.3)} & \textbf{49.2} {\scriptsize (11.3)} & 42.7 {\scriptsize (16.8)} & 39.7 {\scriptsize (22.9)} \\
DINOv3 ViT-Small
& 64.3 {\scriptsize (25.4)} & 76.2 {\scriptsize (20.0)} & \textbf{88.4} {\scriptsize (4.1)} & \underline{86.5} {\scriptsize (3.4)} 
& 30.3 {\scriptsize (21.3)} & 41.0 {\scriptsize (16.5)} & \underline{47.9} {\scriptsize (6.2)} & 44.6 {\scriptsize (4.9)} \\
BioCLIP
& 76.5 {\scriptsize (0.0)} & 61.3 {\scriptsize (34.1)} & 76.5 {\scriptsize (0.0)} & 76.5 {\scriptsize (0.0)} 
& 0.0 {\scriptsize (0.0)} & 3.8 {\scriptsize (8.4)} & 0.0 {\scriptsize (0.0)} & 0.0 {\scriptsize (0.0)} \\

\midrule
\multicolumn{9}{l}{\textbf{PhenoBench}} \\
CLIP ViT-B-16
& 0.0 {\scriptsize (0.0)} & 0.0 {\scriptsize (0.0)} & 11.7 {\scriptsize (24.5)} & 0.9 {\scriptsize (0.9)} 
& 5.1 {\scriptsize (0.0)} & 5.1 {\scriptsize (0.0)} & 5.1 {\scriptsize (0.1)} & 5.0 {\scriptsize (0.1)} \\
CLIP ViT-B-32 L.2b
& 61.1 {\scriptsize (10.0)} & \underline{63.2} {\scriptsize (4.8)} & 62.7 {\scriptsize (8.3)} & \textbf{66.5} {\scriptsize (1.1)} 
& 2.8 {\scriptsize (2.9)} & 3.6 {\scriptsize (2.6)} & 2.9 {\scriptsize (1.9)} & 1.7 {\scriptsize (1.3)} \\
CLIP ViT-B-32
& 9.5 {\scriptsize (5.7)} & 22.3 {\scriptsize (20.1)} & 19.8 {\scriptsize (15.6)} & 49.5 {\scriptsize (14.4)} 
& 3.4 {\scriptsize (0.9)} & 2.9 {\scriptsize (1.7)} & 3.1 {\scriptsize (1.2)} & 2.4 {\scriptsize (1.3)} \\
DINOv3 ConvN.-T.
& 31.0 {\scriptsize (20.9)} & 27.7 {\scriptsize (3.9)} & 32.3 {\scriptsize (4.6)} & 50.0 {\scriptsize (14.3)} 
& 8.2 {\scriptsize (2.7)} & 11.5 {\scriptsize (2.0)} & \underline{13.3} {\scriptsize (2.0)} & \textbf{18.1} {\scriptsize (3.7)} \\
DINOv3 ViT-Small
& 6.3 {\scriptsize (8.0)} & 6.3 {\scriptsize (7.9)} & 11.2 {\scriptsize (11.8)} & 32.6 {\scriptsize (5.9)} 
& 4.4 {\scriptsize (2.7)} & 4.4 {\scriptsize (2.8)} & 5.9 {\scriptsize (3.9)} & 7.4 {\scriptsize (2.7)} \\
BioCLIP
& 62.6 {\scriptsize (6.7)} & 39.1 {\scriptsize (35.9)} & 48.1 {\scriptsize (28.0)} & 50.9 {\scriptsize (28.6)} 
& 0.6 {\scriptsize (0.5)} & 2.5 {\scriptsize (2.5)} & 2.4 {\scriptsize (1.8)} & 1.9 {\scriptsize (2.0)} \\

\midrule
\multicolumn{9}{l}{\textbf{SteelDS - Extended}} \\
CLIP ViT-B-16
& 56.5 {\scriptsize (24.0)} & 32.3 {\scriptsize (25.7)} & 39.6 {\scriptsize (21.7)} & 72.9 {\scriptsize (4.7)} 
& 9.2 {\scriptsize (8.7)} & 14.7 {\scriptsize (7.9)} & 16.0 {\scriptsize (8.7)} & 3.0 {\scriptsize (6.1)} \\
CLIP ViT-B-32 L.2b
& 49.0 {\scriptsize (9.7)} & 53.3 {\scriptsize (4.4)} & 60.1 {\scriptsize (3.9)} & 57.6 {\scriptsize (1.2)} 
& 16.7 {\scriptsize (2.1)} & 17.3 {\scriptsize (1.2)} & 14.1 {\scriptsize (3.7)} & 15.7 {\scriptsize (4.2)} \\
CLIP ViT-B-32
& 68.5 {\scriptsize (5.7)} & 66.2 {\scriptsize (4.8)} & 69.6 {\scriptsize (5.1)} & 73.4 {\scriptsize (1.9)} 
& 7.1 {\scriptsize (6.3)} & 8.8 {\scriptsize (4.3)} & 5.3 {\scriptsize (4.3)} & 3.4 {\scriptsize (3.3)} \\
DINOv3 ConvN.-T.
& 36.4 {\scriptsize (20.6)} & 60.8 {\scriptsize (16.2)} & 64.8 {\scriptsize (16.6)} & 84.0 {\scriptsize (7.1)} 
& 23.6 {\scriptsize (8.3)} & 27.7 {\scriptsize (22.7)} & 39.6 {\scriptsize (12.4)} & 55.2 {\scriptsize (7.8)} \\
DINOv3 ViT-Small
& 63.8 {\scriptsize (40.7)} & 81.8 {\scriptsize (21.1)} & \textbf{88.3} {\scriptsize (3.9)} & \underline{88.0} {\scriptsize (4.6)} 
& 50.5 {\scriptsize (30.1)} & 60.3 {\scriptsize (25.3)} & \textbf{63.5} {\scriptsize (10.1)} & \underline{61.7} {\scriptsize (9.7)} \\
BioCLIP
& 75.6 {\scriptsize (0.1)} & 75.7 {\scriptsize (0.0)} & 75.7 {\scriptsize (0.0)} & 75.7 {\scriptsize (0.0)} 
& 0.0 {\scriptsize (0.1)} & 0.0 {\scriptsize (0.0)} & 0.0 {\scriptsize (0.0)} & 0.0 {\scriptsize (0.0)} \\

\bottomrule
\end{tabular}
\label{tab:iou_evaluation}
\end{table*}

In the industrial domain (SteelDS and SteelDS Extended), DINOv3 performs best across variations of cluster parameters. 
In the agricultural domain (PhenoBench), which previously proved to be challenging, we observe an overall performance degradation for all methods. 
Here, DINOv3-ConvN. still performs best in the critical anomaly detection task and reasonably well in target detection, while CLIP performs best in the easier target detection task but fails in anomaly detection.

Compared to reconstructed baseline and WaW in Table~\ref{tab:av_exp1_miou}, our approach outperforms the baseline in target segmentation on all datasets. 
It also surpasses both our baseline and WaW in the critical anomaly detection task on the challenging PhenoBench dataset by a large margin. 
On SteelDS anomaly detection and PhenoBench target segmentation, we achieve results comparable to those of our baseline and WaW.

Considering the training time of the reconstructed baseline, including WaW, shown in Table~\ref{tab:av_exp1_time}, we conclude that our ViT patch aggregation strategy performs better and faster in the critical anomaly detection task, while achieving comparable performance in target detection. 
Compared to the domain-specific pipeline, which relies on domain-specific knowledge for segment extraction, we trade a slight performance degradation of 2\% IoU for a significant improvement in speed and generalizability through our language-conditioned segmentation pipeline.


\section{CONCLUSION}



This work presented PASTA, a novel approach for weakly supervised anomaly and target detection and segmentation in industrial and agricultural applications.
Our method requires only two sets of images, with and without the target object, to identify and segment anomalies. 
Compared to domain-specific baselines, our approach achieves significantly better segmentation and runtime performance in critical anomaly segmentation tasks, while still delivering comparable results in target segmentation using a domain-agnostic framework, as opposed to slower, domain-specific pipelines.

Our ablation studies show that the choice of clustering hyperparameters affects anomaly detection accuracy differently across various VLM embedding spaces. 
Comparisons with domain-specific baselines further demonstrate that incorporating domain knowledge provides certain benefits, but at the expense of generalization and runtime efficiency.

We also observe that introducing a dedicated segmentation stage significantly improves segmentation quality, albeit with increased inference time. 
In future work, we plan to replace the dedicated segmentation stage with a pixel-wise clustering approach to achieve comparably sharp segmentation performance while reducing inference time.

\section*{ACKNOWLEDGMENT}
The project "KIRAMET KI based Recycling Metalcompound-Waste" (Project number FO999899661) is funded by the Austrian Research Promotion Agency (FFG) and the Federal Ministry for Climate Action, Environment, Energy, Mobility, Innovation, and Technology.
Video and photo material of steel waste were recorded at the Digital Waste Research Lab of the Chair of Waste Processing Technology and Waste Management, TU Leoben. Scholz Austria GmbH contributed as a partner for scrap test specimens and a research collaborator.


\bibliographystyle{IEEEtran}  
\bibliography{main}           

\end{document}